\documentclass[letterpaper, 10pt, conference]{ieeeconf}

\IEEEoverridecommandlockouts                              

\usepackage[pdftex]{graphicx}
\graphicspath{{/figures/}}
\DeclareGraphicsExtensions{.pdf,.jpeg,.png}

\renewcommand{\baselinestretch}{1} 

\usepackage{mathtools} 

\usepackage{cite} 

\usepackage{multirow} 
\usepackage{array} 
\usepackage{makecell} 
\usepackage{tabularx,booktabs} 

\usepackage{graphicx} 
\usepackage[table,xcdraw]{xcolor} 
\usepackage{psfrag} 
\usepackage{color} 

\usepackage{lipsum} 
\usepackage{float} 
\usepackage[hidelinks]{hyperref}  
\usepackage{cleveref} 
\usepackage{siunitx} 
\usepackage[export]{adjustbox} 
\usepackage{textcomp} 

\usepackage{algorithm} 
\usepackage{algpseudocode} 

\usepackage{balance} 
\usepackage{bm} 

\usepackage[compact]{titlesec}
\titlespacing*{\section}{0pt}{*0.8}{*0.8}
\titlespacing*{\subsection}{0pt}{*0.7}{*0.7}
\titlespacing*{\subsubsection}{0pt}{*0.6}{*0.6}

\usepackage[font=small,skip=5pt]{caption}

\usepackage{placeins}

\begin{document}

\title{\LARGE \bf  Stretchable Electrohydraulic Artificial Muscle for Full Motion Ranges in Musculoskeletal Antagonistic Joints}

\author{Amirhossein Kazemipour$^{1}$,
    Ronan Hinchet$^{1}$,
    Robert K. Katzschmann$^{1}$
\thanks{$^{1}$Soft Robotics Lab, ETH Zurich, Switzerland}%
    \thanks{{\tt\footnotesize \{\href{mailto:akazemi@ethz.ch}{akazemi}, \href{mailto:rhinchet@ethz.ch}{rhinchet}, \href{mailto:rkk@ethz.ch}{rkk}\}@ethz.ch}}}



\maketitle
\thispagestyle{empty}
\pagestyle{empty}

\begin{abstract}
Artificial muscles play a crucial role in musculoskeletal robotics and prosthetics to approximate the force-generating functionality of biological muscle. However, current artificial muscle systems are typically limited to either contraction or extension, not both. This limitation hinders the development of fully functional artificial musculoskeletal systems. We address this challenge by introducing an artificial antagonistic muscle system capable of both contraction and extension. Our design integrates non-stretchable electrohydraulic soft actuators (HASELs) with electrostatic clutches within an antagonistic musculoskeletal framework. This configuration enables an antagonistic joint to achieve a full range of motion without displacement loss due to tendon slack. We implement a synchronization method to coordinate muscle and clutch units, ensuring smooth motion profiles and speeds. This approach facilitates seamless transitions between antagonistic muscles at operational frequencies of up to 3.2\,Hz. While our prototype utilizes electrohydraulic actuators, this muscle-clutch concept is adaptable to other non-stretchable artificial muscles, such as McKibben actuators, expanding their capability for extension and full range of motion in antagonistic setups. Our design represents a significant advancement in the development of fundamental components for more functional and efficient artificial musculoskeletal systems, bringing their capabilities closer to those of their biological counterparts.
\end{abstract}


\section{Introduction}
\subsection{Motivation}
\begin{figure}[!ht]
	\centering
    \includegraphics[width=0.95\columnwidth]{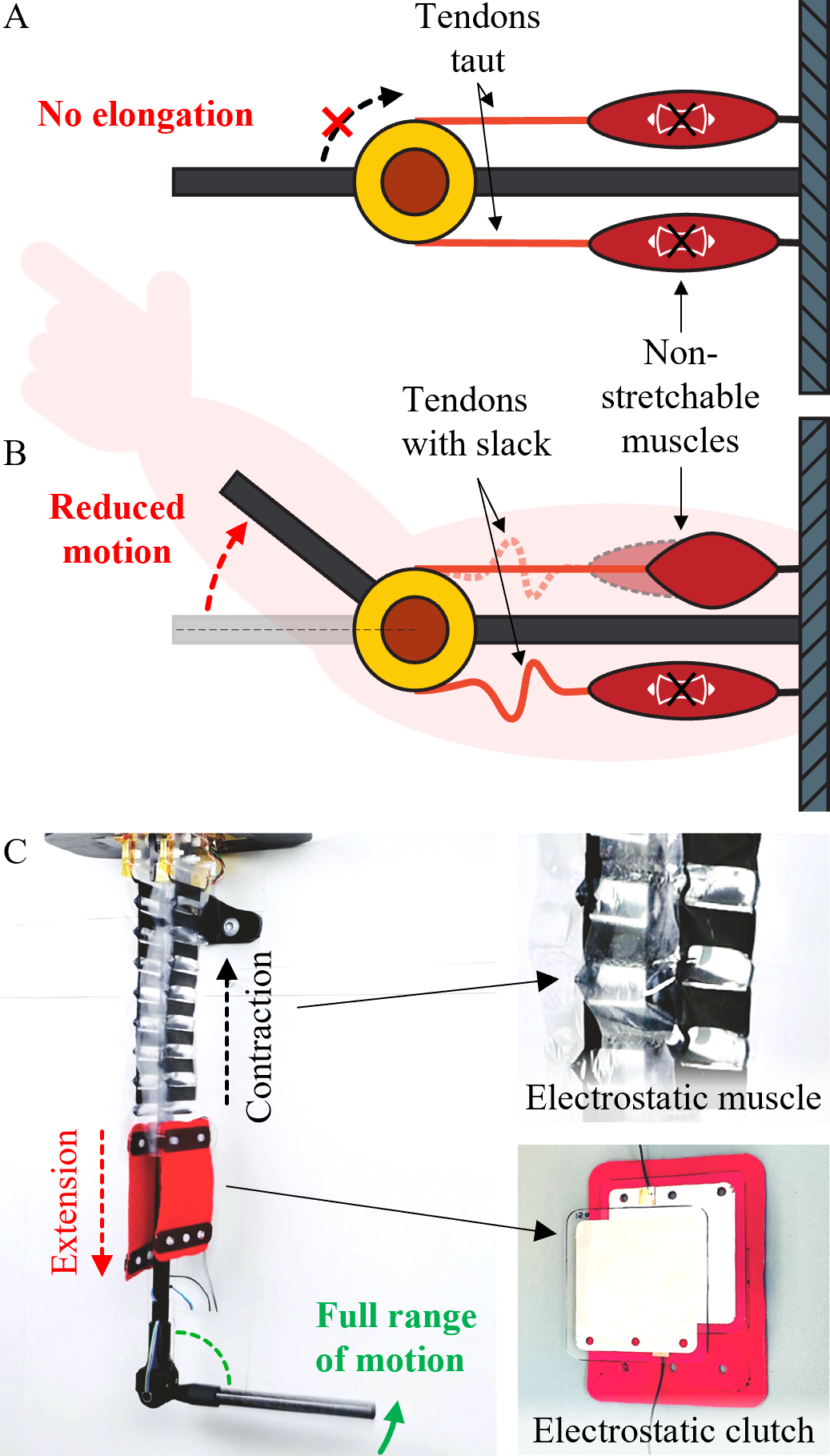}
	\caption{Overview of the proposed artificial muscle system enabling full motion range: (\textbf{A})~In antagonistic setups with non-stretchable muscles, if the tendons are taut, the motion is blocked since these muscles cannot elongate. (\textbf{B})~Adding slack permits movement but reduces the motion range. (\textbf{C})~Our solution uses electrostatic clutches and muscles to enable contraction and extension, restoring full motion range.}
	\label{fig1}
\end{figure}
Advancements in robotics increasingly draw inspiration from the versatility and adaptability of natural organisms. Hybrid rigid-soft robots, which integrate soft materials into rigid structures~\cite{hawkes_soft_2017, wallin_3d_2018, wehner_integrated_2016, byun_electronic_2018}, aim to replicate the seamless movements and dexterity found in nature. Soft elements provide compliance at critical points, such as contact interfaces~\cite{kim_bipedal_2021}, and soft muscles attached via tendons to a rigid skeleton generate motion around passive joints. This design paradigm offers significant advantages over conventional rigid robots~\cite{hutter_anymal_2016, lee_learning_2020, miki_learning_2022}, including reduced power requirements for equivalent torque, enhanced joint accuracy, and a more streamlined form factor~\cite{buchner_electrohydraulic_2024}. By decoupling actuators from the load-bearing structure and joints, musculoskeletal robots also achieve greater payload capacities compared to entirely soft robots.
 
The performance of musculoskeletal robots hinges on both skeletal design and the efficacy of their artificial muscles. Revolute joints, or hinges, are the most prevalent in articulated robots, providing a uniaxial degree of freedom as seen in human finger and toe joints. More complex joints like knees and elbows permit slight rotations or shifts, adding to mechanical complexity. Soft muscles, often arranged in antagonistic pairs akin to biological biceps and triceps, actuate these joints. To mimic this biological functionality, musculoskeletal robots employ both soft fluidic~\cite{hitzmann_anthropomorphic_2018, niiyama_biomechanical_2012, ikemoto_shoulder_2015, kurumaya_musculoskeletal_2016} and electromagnetic~\cite{asano_human_2016, jantsch_anthrob_2013, richter_musculoskeletal_2016} actuators.

Pneumatic Artificial Muscles (PAMs), such as McKibben muscles, were among the earliest soft artificial muscles. Comprising an inflatable bladder within a non-stretchable shell, they contract upon inflation. PAMs are still prevalent in complex bio-inspired robotic systems like the Shadow Hand C3~\cite{cui_pneumatic_2017}, offering excellent strength-to-weight ratios. However, they are limited by slow response times and, in untethered systems, the cumbersome requirement for compressors, tanks, and valves.

To overcome these limitations, various types of soft artificial muscles have been developed~\cite{el-atab_soft_2020}. Notably, Hydraulically Amplified Self-healing Electrostatic (HASEL) actuators~\cite{acome_hydraulically_2018} have emerged as a promising alternative by combining soft fluidic actuation with electrostatic forces. These actuators consist of oil-filled pouches partially covered with electrodes; when activated, the electrodes compress and redistribute the oil, causing the pouch to inflate and contract. Moreover, recent advances have integrated high-frequency self-sensing capabilities into HASELs for real-time displacement estimation and proprioceptive feedback control~\cite{vogt_high-frequency_2024,christoph_self-sensing_2024}, enabling actuation at elevated frequencies without external sensors. Using Polyvinylidene Difluoride (PVDF) terpolymer substrates further reduces the actuation voltage by fivefold while maintaining force and strain performance~\cite{gravert_low-voltage_2024}. However, a substantial drawback remains: these thermoplastic substrates render HASELs non-stretchable, akin to McKibben muscles.

\subsection{Problem Statement}
An important design limitation of musculoskeletal robots is the use of non-stretchable artificial muscles, such as McKibben and HASEL actuators, in antagonistic configurations (e.g., on either side of a hinge). In the human arm, the biceps (agonist) contracts to move the arm, while the triceps (antagonist) relaxes to allow movement. However, in robots, non-stretchable muscles cannot relax and elongate the antagonist, preventing the agonist's contraction and blocking arm motion (Fig.~\ref{fig1}A).

One solution is to introduce slack in the antagonist's tendon, allowing the agonist to contract and bend the arm (Fig.~\ref{fig1}B). However, to straighten the arm back, the slack must be limited to a maximum of half of the muscle contraction amplitude, restricting effective muscle contraction to $50\%$. This significantly limits the motion range and introduces latency, hindering the robot's performance.

Alternatively, incorporating linear clutches in series and synchronized with the muscles can utilize the full muscle contraction. Clutches dynamically block or couple motion, enabling variable stiffness systems. When put in series with muscles, they can disengage the triceps during biceps contraction to bend the arm and vice versa to straighten it back, allowing $100\%$ utilization of muscle contraction. This approach, however, depends on clutch size, weight, holding force, elasticity, and power efficiency.

Existing clutches~\cite{plooij_lock_2015} are mostly rigid. While pneumatic and vacuum jamming clutches can be soft, they require bulky pumps and valves. Electromagnetic clutches, though popular and efficient, are heavy and power-consuming. Mechanical latches reduce power usage but are slow and complex. Magnetorheological fluid designs~\cite{wang_novel_2013} are simpler but heavier, and piezoelectric clutches~\cite{spanner_piezoelectric_2016} are lighter and more efficient yet rigid and complex. These clutches are unsuitable for musculoskeletal robots due to their lack of compliance, high mass, and complexity. 

The recent development of electrostatic clutches (ESclutches) offers superior force-to-mass ($<10^4$) and force-to-power ratio ($<10^5$) ~\cite{diller_effects_2018}. ESclutches are fast, thin (sub-millimeter), light, compact, and flexible~\cite{hinchet_high_2020}. They consume minimal current, and their holding force is adjustable via driving voltage~\cite{hinchet_glove_2022}. They have been integrated into textile~\cite{hinchet_high_2020} in wearable robotics for haptic feedback~\cite{hinchet_glove_2022} and exoskeletons~\cite{diller_lightweight_2016}. ESclutches are well-suited to complement HASELs in musculoskeletal robots, but such a combination has never been evaluated to actuate antagonistic muscle configurations.

\subsection{Contributions}
In this paper, we introduce a novel contractile artificial muscle system capable of stretching by: (1) combining HASEL actuators with ESclutches in series; (2) integrating this hybrid system into a musculoskeletal structure with antagonistic muscles; and (3) demonstrating the synergistic operation of HASELs and ESclutches to enhance limb motion (Fig.~\ref{fig1}C).

The combination of HASEL actuators and ESclutches is advantageous as both are thin, flexible, and based on electrostatics, enabling fast and efficient actuation using high voltages and low currents for safety. This facilitates seamless integration and control, with driving circuits that can be combined and miniaturized to a few cubic centimeters~\cite{gravert_low-voltage_2024}. This work presents the first evaluation and demonstration of such a hybrid actuation system in a musculoskeletal framework.

\section{METHODOLOGY}
This section details our solution, covering theoretical analysis, design and characterization of the clutch and HASEL actuator, their integration into a unified muscle capable of both extension and contraction, incorporation into an antagonistic joint musculoskeletal system, and control methods for smooth operation.

\subsection{Theoretical design analysis}
We analyze an antagonistic joint actuated by HASEL muscle packs integrated with ESclutches. By placing an ESclutch in series with each HASEL, tendon slack (set to \(50\%\) of the HASEL strain \(\sigma\) at force \(F_\mathrm{h}\)) is eliminated. This leads to two design strategies:

\begin{enumerate}
    \item \textit{Compact Configuration:} Reduce the effective length of the HASEL by half,
    \begin{equation}
        L_\mathrm{h}' = 0.5\,L_\mathrm{h},
        \label{eq:equation2}
    \end{equation}
    which—neglecting resisting forces—achieves the same absolute range of motion as a full-length HASEL operating without clutches.
    
    \item \textit{Enhanced Range of Motion:} Retain the full length of the HASEL,
    \begin{equation}
        L_\mathrm{h}' = L_\mathrm{h},
    \end{equation}
    allowing for larger absolute displacement.
\end{enumerate}

For both cases, we consider a HASEL with original length \(L_\mathrm{h}\) and width \(W_\mathrm{h}\), paired with an ESclutch of matching width (\textit{i.e.}, \(W_\mathrm{c}=W_\mathrm{h}\)). The ESclutch is characterized by a surface friction force density \(P_\mathrm{c}\) (measured experimentally). Thus, the theoretical clutch length needed to transmit the HASEL force is:
\begin{equation}
    L_\mathrm{c} = \frac{F_\mathrm{h}}{P_\mathrm{c}\,W_\mathrm{h}},
    \label{eq:equation1}
\end{equation}
where \(F_\mathrm{h}\) is the HASEL operating force. In practice, \(L_\mathrm{c}\) is slightly oversized to accommodate packaging and enhance the blocking force.

Since the ESclutch is packaged with a soft elastic textile (of thickness \(T_\mathrm{t}\) and modulus \(E_\mathrm{t}\)), an additional resisting force arises:
\begin{equation}
    F_\mathrm{t} = \sigma\,\frac{L_\mathrm{h}'}{L_\mathrm{c}}\,E_\mathrm{t}\,W_\mathrm{h}\,T_\mathrm{t},
    \label{eq:equation3}
\end{equation}
with \(\sigma\) being the HASEL strain at the operating force. This extra load must be overcome in both configurations. To ensure sufficient force output, the HASEL width may be increased as:
\begin{equation}
    W_\mathrm{h}' = W_\mathrm{h} + \frac{F_\mathrm{t}\,W_\mathrm{h}}{F_\mathrm{h}},
    \label{eq:equation4}
\end{equation}
or, if the width remains unchanged, the HASEL will operate at a slightly reduced strain.

In summary, our ESclutch dimensions are derived from these equations using experimental values for \(F_\mathrm{h}\), \(P_\mathrm{c}\), \(T_\mathrm{t}\), and \(E_\mathrm{t}\). We slightly oversize it beyond the theoretical minimums to ensure robust blocking force and high-speed switching. Thus, whether adopting the compact (half-length) or enhanced (full-length) configuration, integrating clutches enables full utilization of the actuator’s range—provided that the extra elastic resistance is properly managed.

\subsection{Electrostatic Clutch Characterization}
\begin{figure}[thpb]
	\centering
 	\includegraphics[width=1\columnwidth]{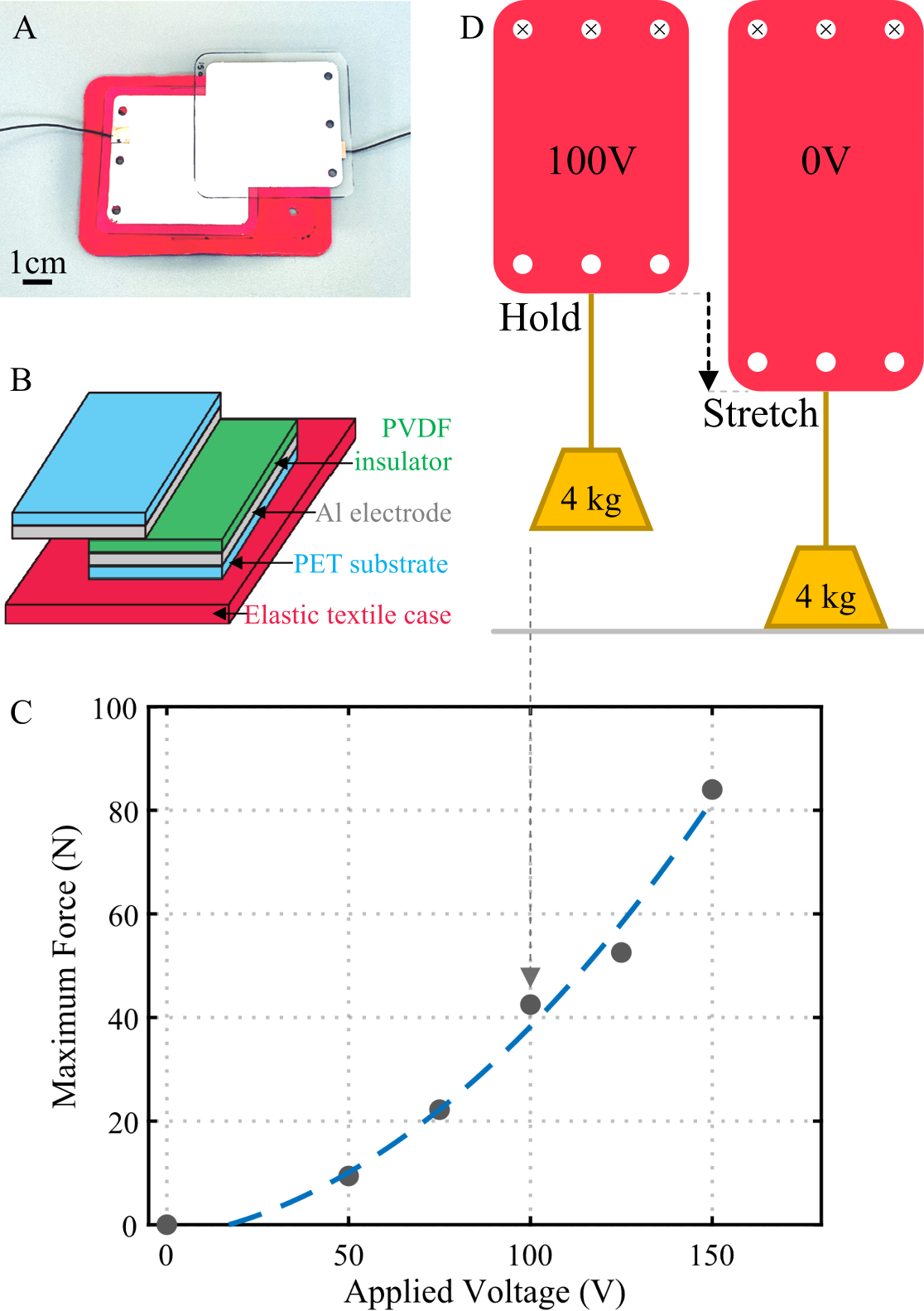}
	\caption{Overview of the ESclutch design and force range. (\textbf{A}) Photo of the ESclutch showing its thin and compact format. (\textbf{B}) Structure of the ESclutch describing its different layers. (\textbf{C}) Characterization of the maximum holding force of the ESclutch as a function of the voltage applied up to \SI{150}{\volt} with a \SI{10}{\hertz} AC square signal, which shows that it can sustain \SI{8.41}{\kilo\gram}. (\textbf{D}) Illustration of the ESclutch behavior under actuation.}
	\label{fig2}
\end{figure}

ESclutches were designed and fabricated following the process described in~\cite{hinchet_glove_2022}. Each ESclutch consists of two \SI{5}{\centi\meter} by \SI{5}{\centi\meter} electrodes that can slide on each other (Fig.~\ref{fig2}A). They are composed of \SI{125}{\micro\meter} thick PET films with \SI{50}{\nano\meter} thick Al electrodes on top. The electrodes are separated by a \SI{6}{\micro\meter} thick PVDF terpolymer film deposited on one of them (Fig.~\ref{fig2}B). Electric wires are connected at the extremities, and holes are drilled for attachment. Finally, the ESclutch is placed between 2 pieces of soft stretchable textile, having Young's modulus of around \SI{0.1}{\mega\pascal}~\cite{hinchet_glove_2022}, for protection and to ensure return to its original position upon release. The resulting clutch is \SI{6}{\centi\meter} wide and \SI{9}{\centi\meter} long including casing and is \SI{1}{\milli\meter} thick in total for a weight of \SI{4.4}{\gram}. Inside, the electrodes' overlap is \SI{15}{\centi\meter\squared} in its released state, and the clutch can stretch up to $44\%$ or \SI{4}{\centi\meter}.

ESclutches were characterized by applying a symmetric square voltage at \SI{10}{\hertz} between the electrodes and then pulling on the attached clutch using a force sensor until the electrodes slide. This gave us the maximum holding force of the clutch depending on the applied voltage (Fig. \ref{fig2}C). It shows that our clutch can hold \SI{4.25}{\kilo\gram} when actuated at \SI{100}{\volt} (Fig.~\ref{fig2}D) which is equivalent to a shear stress of \SI{2.8}{\newton\per\square\centi\meter}. But the clutch can easily block \SI{8.41}{\kilo\gram} at \SI{150}{\volt} if needed for a bigger robot. This corresponds to a shear stress of \SI{5.6}{\newton\per\square\centi\meter}, which is on the same order of magnitude as~\cite{hinchet_high_2020}. Such a clutch can usually lock in \SI{5}{\milli\second} and release in \SI{15}{\milli\second}.

\subsection{HASEL Characterization}
\begin{figure}[thpb]
	\centering
 	\includegraphics[width=0.95\columnwidth]{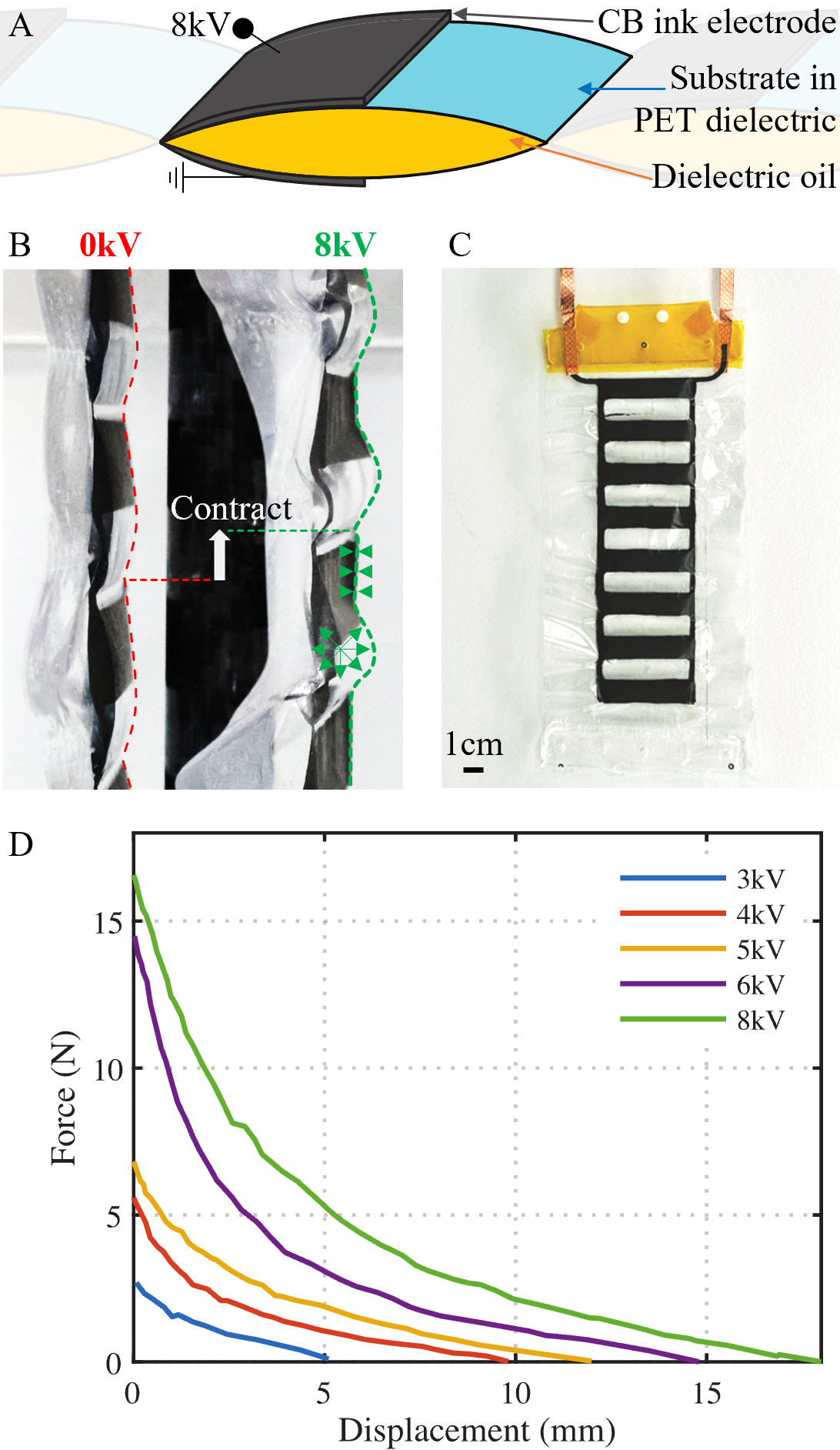}
	\caption{Overview of the HASEL actuator design and force range. (\textbf{A}) Structure of one HASEL pouch describing its constitutive layers and (\textbf{B}) zoom-in photo of HASELs at the relaxed and contracted state, which show the bulging and contraction of the pouch when actuated at \SI{8}{\kilo\volt} with a DC square signal. (\textbf{C}) Photo of one HASEL pack of 8 pouches in series. (\textbf{D}) The force-displacement characterization curves of the HASEL depend on the applied voltage and show that the actuator can generate a force up to \SI{16.3}{\newton} or a displacement up to \SI{18.0}{\milli\meter} when actuated at \SI{8}{\kilo\volt}.}
	\label{fig3}
\end{figure}

Thermoplastic-based HASEL actuators consist of a non-stretchable yet flexible shell encapsulating a liquid dielectric, flanked by electrodes (Fig.\ref{fig3}A). When voltage is applied, the electrodes attract, squeezing the liquid and redistributing it within the actuator (Fig.\ref{fig3}B). As the pouch deforms, the actuator contracts.

HASELs were fabricated following the process in~\cite{mitchell_easy--implement_2019}. Each pouch measures \SI{4.5}{\centi\meter} wide and \SI{2}{\centi\meter} long, with 50\% electrode coverage. The shell is a \SI{15}{\micro\meter} thick heat-sealable Mylar sheet. Black carbon inks serve as electrodes, with silicone oil as the dielectric fluid. The HASEL muscle comprises 8 pouches (Fig.~\ref{fig3}C), totaling \SI{16}{\centi\meter} in length and weighing \SI{13.7}{\gram}.

HASEL actuators exhibit nonlinear force-displacement curves (Fig.~\ref{fig3}D), obtained by anchoring one end to a load cell, applying force to the other, and tracking motion with a laser sensor. HASELs show higher displacement at lower loads and vice versa. This characteristic is crucial for robotic system design, requiring consideration of the specific force-strain curve to meet load and displacement targets. HASEL performance can be customized by adjusting geometry, structure, and materials.

\subsection{System Integration}
\begin{figure}[th]
	\centering
	\includegraphics[width=0.85\columnwidth]{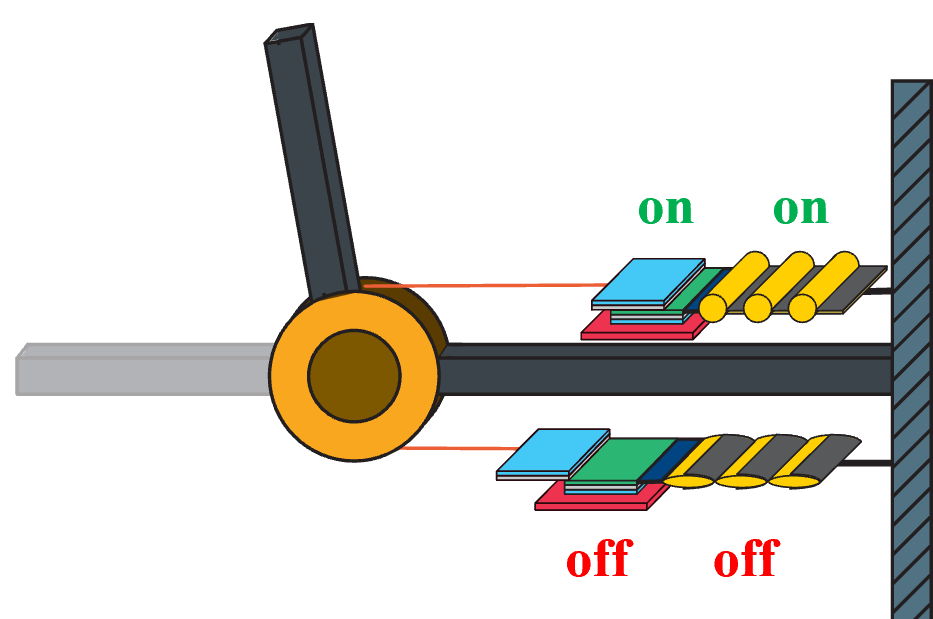}  
	\caption{Musculoskeletal robotic joint actuated with the HASEL-clutch units. When the bottom HASEL-clutch unit is in the 'Off-Off' state, it can elongate, and by setting the top HASEL-clutch unit to the 'On-On' state, the limb moves upward. For downward motion, the operation is reversed.}
	\label{fig4}
\end{figure}

An articulated limb was constructed by connecting a carbon fiber tube with a 3D-printed PLA joint and ball bearings to a \SI{12}{\centi\meter}-long tube weighing \SI{3.8}{\gram}. On the limb, electrostatic HASEL-clutch units were connected in series and arranged antagonistically (Fig.~\ref{fig4}). Fishing line tendons were attached to the ends of the electrostatic HASEL-clutch units and were taut when the limb was in the middle position.

For design optimization, we considered the fabricated HASELs of \SI{4.5}{\centi\meter} wide and \SI{16}{\centi\meter} long generating a force of \SI{1}{\newton} at a strain of $8\%$ at \SI{8}{\kilo\volt}; and the fabricated clutches generating a friction force density of \SI{5.5}{\newton\per\square\centi\meter} at \SI{150}{\volt} and packaged with \SI{1}{\milli\meter} thick stretchable textile (elasticity modulus \SI{100}{\kilo\pascal}).

According to \cref{eq:equation1}, a clutch of \SI{0.18}{\square\centi\meter} or \SI{0.4}{\milli\meter} in length and \SI{4.5}{\centi\meter} in width would theoretically suffice to transmit the HASEL force. However, this size would limit stretchability and increase resistance from the antagonistic clutch's elastic sleeve. Therefore, we oversized the clutch to enhance blocking force, reduce strain, and minimize resistance, which is crucial for high-speed switching and minimizing leg resonance. Additionally, the packaging was designed to maintain system modularity.

\subsection{Control Strategy}

\begin{figure}[h]
    \centering
    \includegraphics[width=0.85\columnwidth]{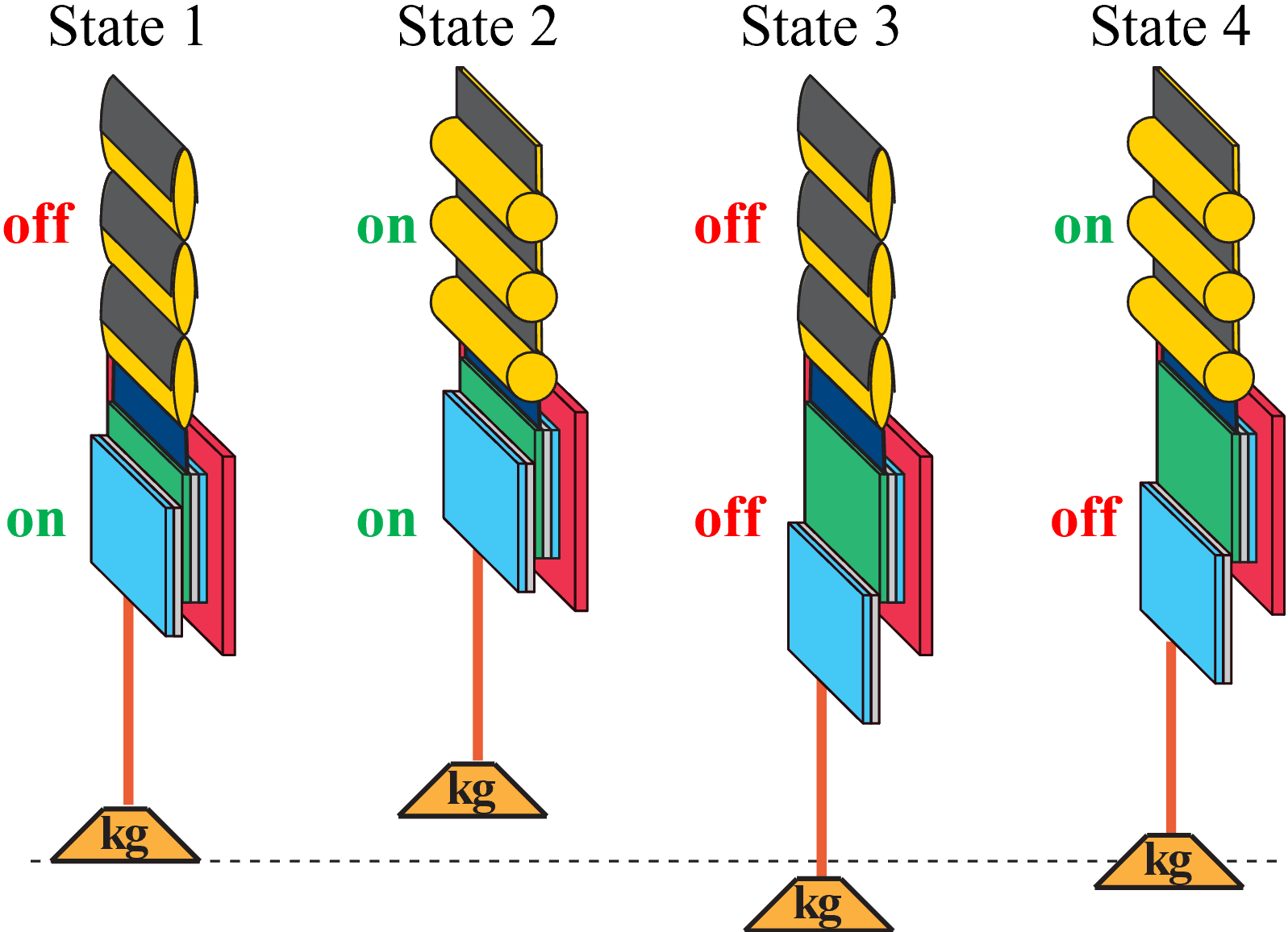}
    \caption{State diagram for HASEL-clutch control. Each HASEL-clutch pair can be in one of four states: State~1 (on/off), State~2 (on/on), State~3 (off/off), and State~4 (off/on). Transitions between states enable coordinated limb movements and braking.}
    \label{fig5}
\end{figure}

Our system is actuated using four control signals: two for HASEL actuators and two for clutches. Achieving a full range of motion requires synchronizing these signals through a state machine logic with a feedforward control strategy. Clutches are driven by an AC symmetric square wave at \SI{10}{\hertz} to optimize the release speed and the locking force; note that the clutches are evaluated up to \SI{100}{\volt} because, in practice, this was sufficient to provide the necessary locking force while minimizing the risk of electrical breakdown. For HASELs, we use ramp signals to control their displacement, ensuring smooth, gradual motion without abrupt transitions. The linear lever arm design ensures that HASEL displacement directly translates to joint angles.

Synchronization between clutches and HASELs is essential for effective musculoskeletal operation. When a clutch is "off," it allows stretching; when "on," it locks the limb. Similarly, a HASEL in the "on" state contracts, and in the "off" state, it releases contraction. Figure~\ref{fig5} illustrates the possible states of a HASEL-clutch pair.

To move the limb rightward, the left HASEL-clutch pair is set to State~3 (off/off) and the right pair to State~2 (on/on), causing the right HASEL to contract and its clutch to lock, thereby moving the limb right. Leftward movement is achieved by reversing these settings. Additionally, engaging the opposite clutch allows for rapid stopping by controlling the braking force through the ESclutch's voltage-dependent friction~\cite{hinchet_glove_2022}. This enables variable stiffness and impedance in the joint, enhancing control speed and responsiveness.

\section{Results and Discussions}

\subsection{Experimental Setup}
The setup included four high-voltage amplifiers (one Trek 20/20C, one Trek 610E, two PolyK PK-HVA10005) to power clutches and HASEL actuators. An RLS-RM08 magnetic encoder measured joint angles, and a NI DAQ 6343-USB acquired data. MATLAB with the data acquisition toolbox controlled the system, sending commands to amplifiers and recording voltages and joint angles via the DAQ.

\subsection{Experimental Validation}
\begin{figure}[h]
	\centering
	\includegraphics[width=1\columnwidth]{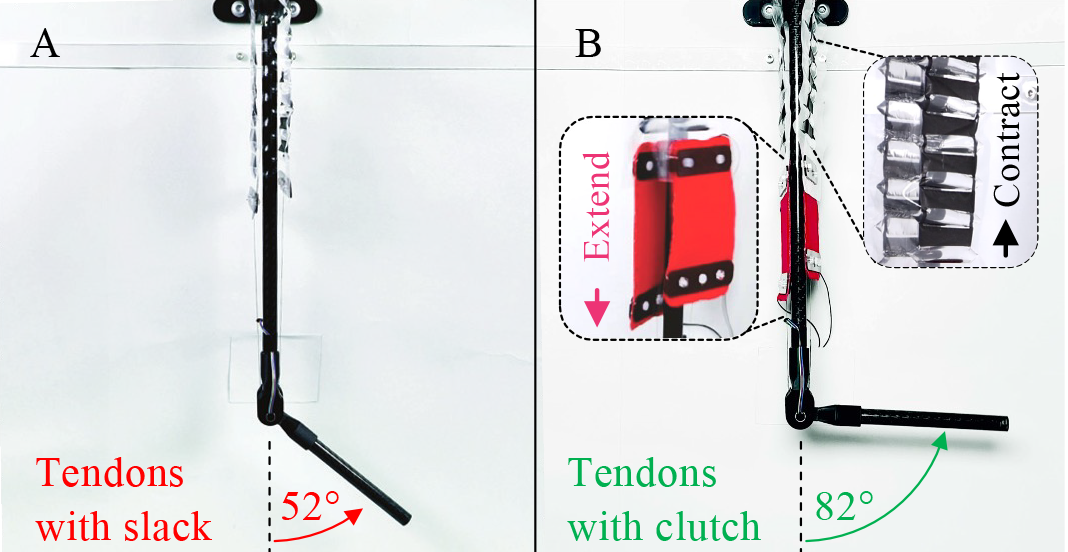}  
	\caption{Comparison of joint range of motion achieved with (\textbf{A}) a configuration using only HASEL actuators for actuation without clutches versus (\textbf{B}) the integrated muscle-clutch mechanism. Despite identical actuation of the HASEL actuators, the muscle-clutch system enables an increased range of motion of $\pm82^\circ$ over the muscle-only arrangement of $\pm52^\circ$.}
	\label{fig6}
\end{figure}

Both muscle mechanisms---HASELs with and without ESclutches---were evaluated on the artificial limb setup (Fig.~\ref{fig6}) using the same voltage command (a ramp signal with a peak of \SI{8}{\kilo\volt}) and frequency (\SI{2.5}{\hertz}), as well as identical tendon anchor points. With only HASELs and optimized tendon slack, the limb rotated \SI{52}{\degree} (Fig.\ref{fig6}A). In contrast, combining ESclutches with the same HASELs increased rotation to \SI{82}{\degree} (Fig.~\ref{fig6}B), resulting in approximately a 58\% increase in usable strain in an antagonistic muscle configuration.

\begin{figure}[thpb]
	\centering
	\includegraphics[width=1\columnwidth]{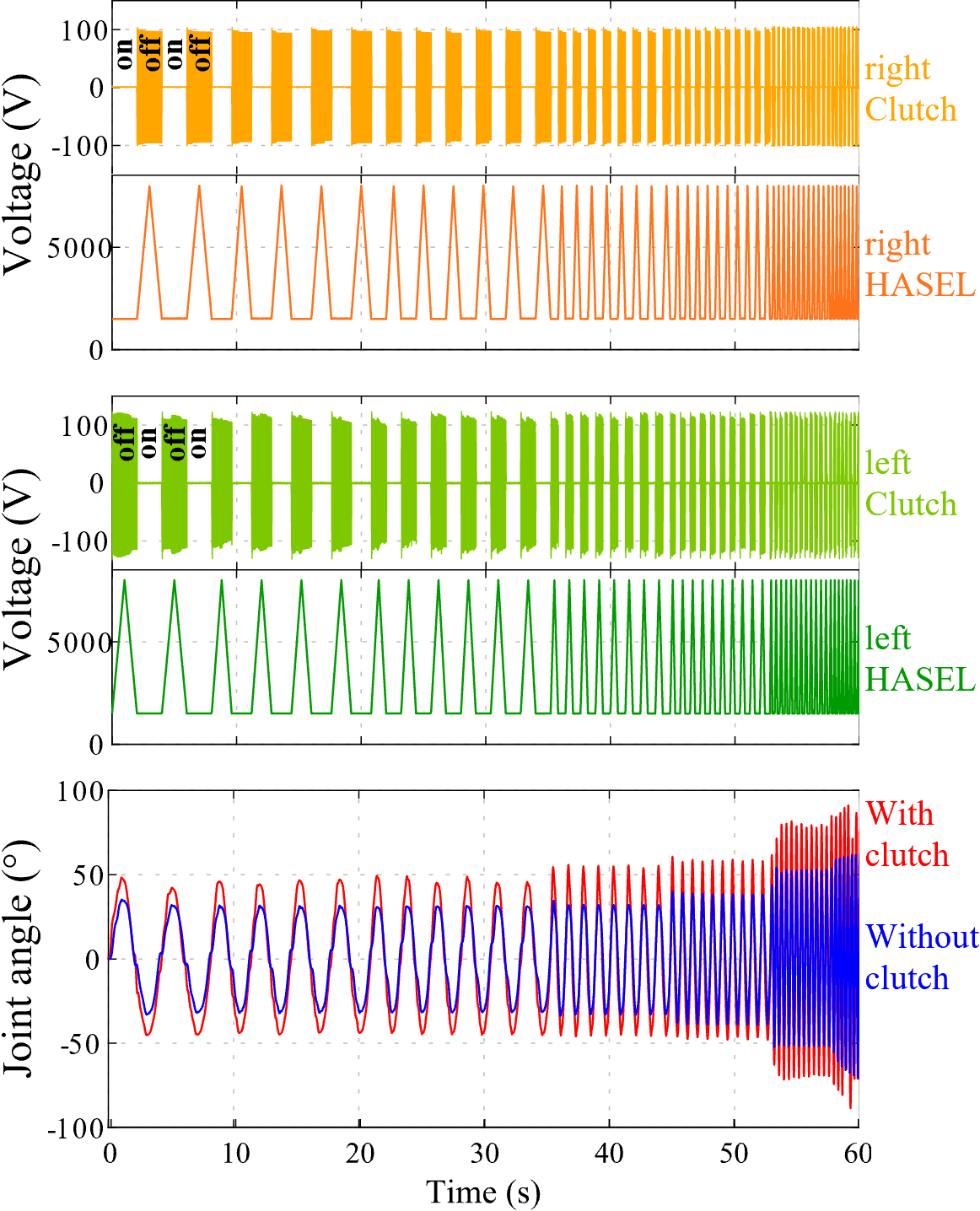}  
	\caption{Experimental results comparing HASEL-only and HASEL-clutch systems. Applied voltages to left/right clutches and HASELs are shown, along with resulting joint angle transitions. The graph demonstrates smooth motion across increasing actuation frequencies, with the HASEL-clutch solution consistently achieving a higher range of motion compared to the HASEL-only system while increasing the actuation frequency up to \SI{3.2}{\hertz}.}
	\label{fig7}
\end{figure}

\begin{figure*}[!ht]
        \centering
        \includegraphics[width=1\textwidth]{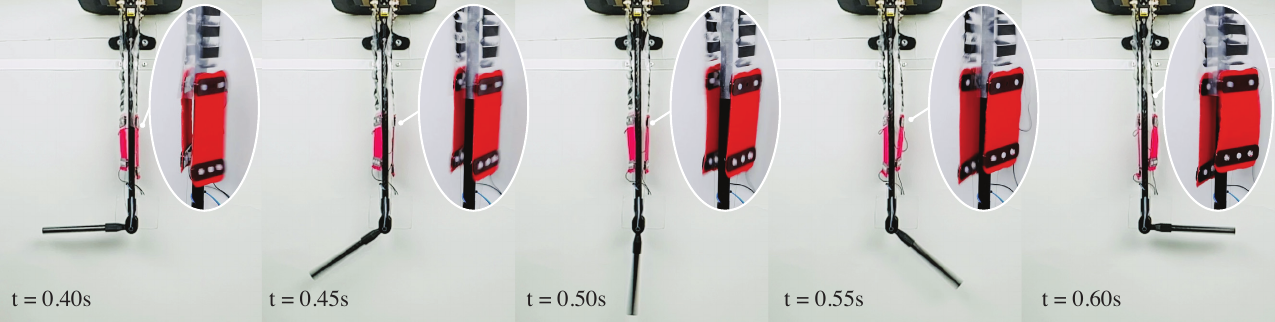}
        \caption{Snapshots of the experiment where the joint is cycling at a freq. of \SI{2.5}{\hertz}, achieving a bidirectional range of motion of $\approx$\SI{160}{\degree}.}
        \label{fig8}
\end{figure*}

The faster actuation speed of ESclutches compared to HASELs facilitates their synchronization, allowing them to be engaged simultaneously. Additionally, since their force is proportional to the applied voltage—and the ESclutch operates at a lower voltage than the HASEL—it is possible to control both using a single voltage source (the high voltage of the HASEL) with a simple voltage divider supplying the ESclutch. This greatly simplifies the control electronics.

We tested the HASEL-clutch system by actuating it at different frequencies. Figure~\ref{fig7} displays the actuation patterns, where we alternately activated the right and left HASELs along with their respective ESclutches. The coordinated voltage patterns, up to \SI{3.2}{\hertz}, demonstrate smooth limb movements without abrupt discontinuities. The resulting joint angle shows seamless transitions between states. Figure~\ref{fig8} presents snapshots of the joint in motion at \SI{2.5}{\hertz}, illustrating rapid, cyclic movements with a wide range of motion.

\begin{figure}[thpb]
	\centering
	\includegraphics[trim= 4mm 0mm 6mm 4mm, width=1\columnwidth,clip]{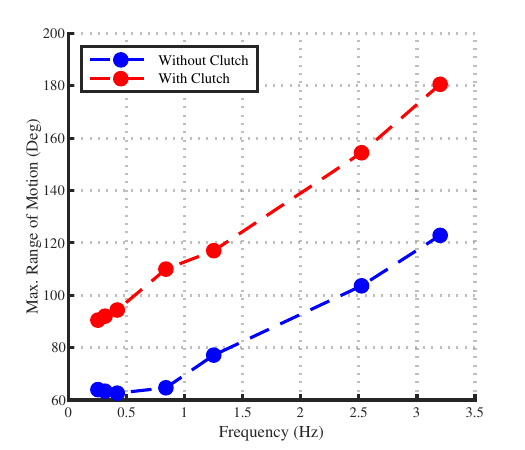}  
	\caption{Maximum range of motion versus actuation frequency for the HASEL-only and clutch-muscle actuated joint. Range of motion expands at higher frequencies due to stretchable clutch sleeves acting as a series elastic component, allowing storage and release of elastic energy to enhance actuation force and motion range. Smooth transitions are observed up to \SI{3.2}{\hertz}.}
	\label{fig9}
\end{figure}

Figure~\ref{fig9} compares the maximum range of motion versus actuation frequency for both HASEL configurations. The HASEL-clutch system consistently achieves a higher range of motion up to \SI{3.2}{\hertz}. As the actuation frequency increased, the maximum range of motion also expanded, reaching up to \SI{180}{\degree}, beyond which joint integrity was compromised—likely due to inertia and resonance phenomena from the clutches' elastic textile sleeves acting like springs. This behavior resembles the series elastic components of muscles, storing and releasing elastic energy to amplify movement amplitude, force, and speed, which can be advantageous when properly controlled.

\section{Conclusion and Future Work}
In this paper, we presented a novel approach to create an artificial muscle that can perform both contraction and extension movements. Our design combines HASEL actuators with ESclutches. We integrated these components into a musculoskeletal system and utilized these HASEL-clutch units to drive an antagonistically actuated joint. This design enabled us to achieve complete functionality of the muscles, ensuring a full range of motion without losing displacement due to slack. 

A limitation of our study is that the limb was intentionally kept  lightweight to simplify the experimental setup and focus on validating the actuation concept. While this design choice streamlined our initial validation, it also lays the groundwork for future studies to explore system performance under heavier loads. Additionally, our vertical setup benefits from gravity aiding the limb’s return to the neutral position; studying the system in a horizontal configuration would offer better insight into the role of the series elastic component introduced by the clutch. Future work could also involve modeling the system to optimize clutch elasticity for specific resonance frequencies, thereby maximizing amplitude and speed, and exploring how the released HASEL could assist in returning the limb through improved synchronization.

Notably, our approach is not restricted to electrostatic actuators alone. It can also be applied to other types of non-stretchable muscles, such as McKibben's muscles \cite{tanaka_back-stretchable_2023,tanaka_fiber_2024} allowing them to extend and achieve full range of motion in an antagonistic setup. This versatility highlights the potential broad applicability of our solution in advancing artificial muscle technology.

\section{Acknowledgments}
This work was supported by the SNSF Project Grant \#200021\_\,215489. This support was instrumental in the development of this research. The authors also thank Piezotech-Arkema for providing the P(VDF-TrFE-CTFE) polymer used to fabricate the ESclutches. 

\addtolength{\textheight}{-4.7cm}   

\begingroup
\footnotesize
\renewcommand{\baselinestretch}{0.95}\normalsize
\bibliographystyle{IEEEtran}
\bibliography{IEEEabrv,refs}
\endgroup
%

\end{document}